\documentclass[acmtog,screen=true]{acmart}

\AtBeginDocument{%
  \providecommand\BibTeX{{%
    \normalfont B\kern-0.5em{\scshape i\kern-0.25em b}\kern-0.8em\TeX}}}

\setcopyright{acmlicensed}
\acmJournal{TOG}
\acmYear{2022}
\acmVolume{41}
\acmNumber{4}
\acmArticle{13}
\acmMonth{7}
\acmDOI{10.1145/3528223.3530173}

\copyrightyear{2022} 
\acmYear{2022} 
\setcopyright{acmcopyright}\acmConference[SIGGRAPH '22 Conference Proceedings]{Special Interest Group on Computer Graphics and Interactive Techniques Conference Proceedings}{August 7--11, 2022}{Vancouver, BC, Canada}
\acmBooktitle{Special Interest Group on Computer Graphics and Interactive Techniques Conference Proceedings (SIGGRAPH '22 Conference Proceedings), August 7--11, 2022, Vancouver, BC, Canada}
\acmPrice{15.00}
\acmDOI{10.1145/3528233.3530707}
\acmISBN{978-1-4503-9337-9/22/08}

\acmSubmissionID{281s1}

\setcitestyle{square}

\usepackage{soul}
\usepackage{microtype}
\usepackage{amsmath}
\usepackage{amsfonts}
\usepackage{booktabs}
\usepackage{hyperref}
\usepackage{multirow}
\usepackage{graphicx}
\usepackage{dsfont}
\usepackage{makecell}
\usepackage{pifont}
\usepackage{xcolor}
\usepackage{wrapfig}
\usepackage{nicefrac}
\usepackage{natbib}
\usepackage{caption}
\usepackage{xspace}
\usepackage{algorithm}
\usepackage{algpseudocode}
\usepackage{tikz}
\usetikzlibrary{spy, shapes, arrows, matrix, pgfplots.groupplots}
\usetikzlibrary{backgrounds}
\usetikzlibrary{arrows,shapes}
\usetikzlibrary{tikzmark}
\usetikzlibrary{calc}
\usepackage{ctable}
\usepackage{xcolor}

\def\myfigure#1#2{%
    \begin{figure}[tb]%
    \centering\includegraphics*[width = \linewidth]{figures/#1}%
    \vspace{-.3cm}%
    \caption{#2}%
    \label{fig:#1}%
    \end{figure}%
}

\def\mycfigure#1#2{%
    \begin{figure*}[htb]%
    \centering\includegraphics*[width = \linewidth]{figures/#1}%
    \vspace{-.2cm}%
    \caption{#2}%
    \label{fig:#1}%
    \end{figure*}%
}

\soulregister\cref7
\soulregister\cite7
\soulregister\cite7
\soulregister\shortcite7
\soulregister\eg0
\soulregister\ie0
\soulregister\etal0

\newcommand{\mysection}[2]{\section{#1}\label{sec:#2}}
\newcommand{\mysubsection}[2]{\subsection{#1}\label{sec:#2}}

\newcommand{\refSec}[1]{Sec.~\ref{sec:#1}}
\newcommand{\refFig}[1]{Fig.~\ref{fig:#1}}
\newcommand{\refEq}[1]{Eq.~\ref{eq:#1}}
\newcommand{\refTab}[1]{Tab.~\ref{tab:#1}}

\newcommand{\ie}{i.e.,\ }
\newcommand{\eg}{e.g.,\ }

\newcommand{\mymath}[2]{\newcommand{#1}{\TextOrMath{$#2$\xspace}{#2}}}

\newcommand{\unsure}[1]{{\sethlcolor{yellow}\hl{#1}}}

\usepackage{titlesec}
\titlespacing*{\paragraph}{0pt}{1.0ex plus 1ex minus .2ex}{1em}

\definecolor{revisedcolor}{rgb}{0,0,0}
\newcommand{\revised}[1]{\textcolor{revisedcolor}{#1}}

\definecolor{colorA}{HTML}{4285f4}
\definecolor{colorB}{HTML}{ea4335}
\definecolor{colorC}{HTML}{fbbc04}
\definecolor{colorD}{HTML}{34a853}
\definecolor{colorE}{HTML}{ff6d01}
\definecolor{colorF}{HTML}{46bdc6}
\definecolor{colorG}{HTML}{d19ff5}
\definecolor{colorH}{HTML}{f59ff5}
\definecolor{colorI}{HTML}{9ff5da}
\definecolor{colorJ}{HTML}{acf59f}
\definecolor{colorK}{HTML}{d70eed}
\definecolor{colorL}{HTML}{811ae8}

\definecolor{colorY}{HTML}{000000}
\definecolor{colorZ}{HTML}{777777}

\newcommand{\method}[1]{\textcolor{color#1}{\texttt{#1}}}
\newcommand{\methodb}[1]{#1}

\colorlet{colorNeRF-PT}{colorA}
\colorlet{colorMLPwPE}{colorL}
\colorlet{colorMLP}{colorA}
\colorlet{colorGrid}{colorB}
\colorlet{colorGridL}{colorB}
\colorlet{colorReLUField}{colorC}
\colorlet{colorReLUFieldL}{colorC}
\colorlet{colorPlenoxels}{colorD}
\colorlet{colorDVGo}{colorE}
\colorlet{colorNeRF-TF}{colorF}
\colorlet{colorRFBig}{colorG}
\colorlet{colorRFLong}{colorH}
\colorlet{colorRFNoPro}{colorI}
\colorlet{colorRFNoDifReg}{colorJ}
\colorlet{colorNGP}{colorK}

\usepackage[nolist]{acronym}
\begin{acronym}
\acro{NN}{Neural Network}
\acro{PSNR}{Peak Signal to Noise Ratio}
\acro{CNN}{Convolutional Neural Network}
\acro{MC}{Monte Carlo}
\acro{NeRF}{Neural Radiance Fields}
\acro{MLP}{Multi-layer Perceptron}
\acro{ReLU}{Rectified Linear Unit}
\acro{SH}{Spherical Harmonics}
\end{acronym}

\newcommand{\scene}[1]{\textsc{#1}}
\def\figurePath{images/}

\def\myfigure#1#2{\begin{figure}[ht]\centering\includegraphics*[width = \linewidth]{\figurePath#1}\vspace{-.2cm}\caption{#2}\label{fig:#1}\end{figure}}

\def\mycfigure#1#2{\begin{figure*}[t]\centering\includegraphics*[clip, width = \linewidth]{\figurePath#1}\vspace{-.2cm}\caption{#2}\label{fig:#1}\end{figure*}}

\soulregister\ref7
\soulregister\cite7
\soulregister\citeetal7
\soulregister\refSec7
\soulregister\refFig7
\soulregister\refTbl7
\soulregister\refEq7
\soulregister\cite7
\soulregister\ref7
\soulregister\pageref7
\soulregister\shortcite7
\soulregister\eg7
\soulregister\ie7
\soulregister\etal7
\soulregister\unsure7

\DeclareGraphicsExtensions{.png,.jpg,.pdf,.ai,.psd}
\newcommand{\name}{ReLU Field\xspace}
\newcommand{\names}{ReLU Fields\xspace}

\begin{document}

\title{\names: The Little Non-linearity That Could}

\author{Animesh Karnewar}
\affiliation{%
	\institution{University College London}
	\country{UK}
}

\author{Tobias Ritschel}
\email{t.ritschel@ucl.ac.uk}
\affiliation{
  \institution{University College London}
  \country{UK}
}

\author{Oliver Wang}
\email{owang@adobe.com}
\affiliation{
  \institution{Adobe Research}
  \country{USA}
}

\author{Niloy J. Mitra}
\email{n.mitra@cs.ucl.ac.uk}
\orcid{0000-0002-2597-0914}
\affiliation{
  \institution{Adobe Research}
\country{USA}
}
\affiliation{
  \institution{University College London}
 \country{UK}
}

\keywords{neural representations, regular data structures, volume rendering, spatial representations}

\begin{teaserfigure}
   \includegraphics[width=\textwidth]{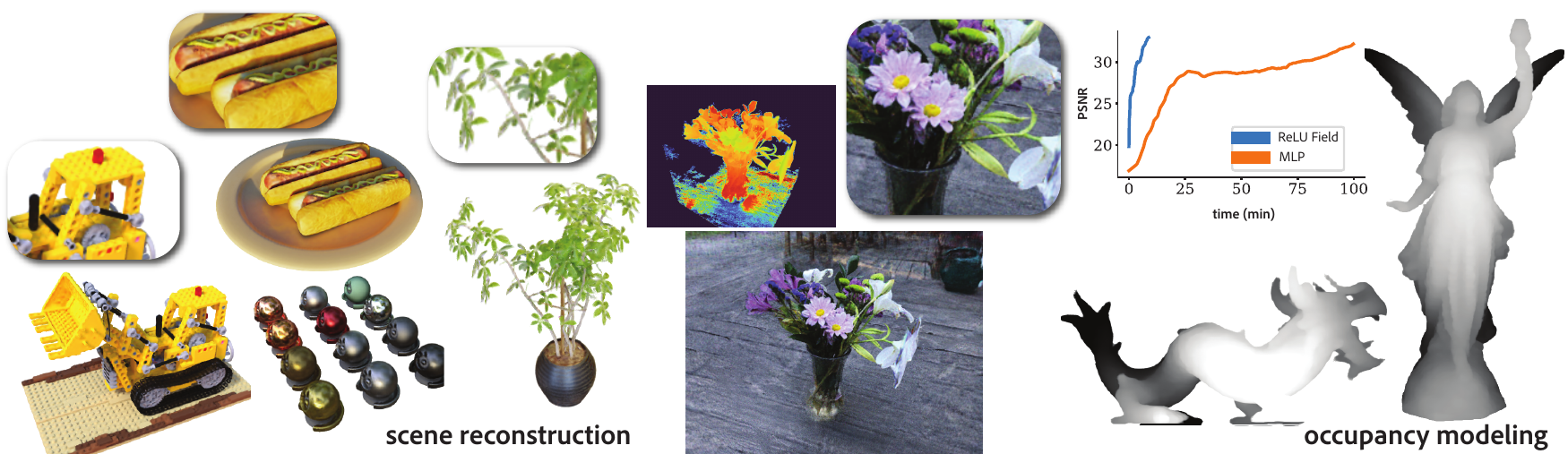}
   \caption{We present a method to represent complex signals such as images or 3D scenes, both volumetric (left) and surface (right), on regularly sampled grid vertices. Our method is able to match the expressiveness of coordinate-based MLPs while retaining reconstruction and rendering speed of voxel grids, \textit{without} requiring any neural networks or sparse data structures.
   As a result it converges significantly faster (inset plot).
   }
   \label{fig:Teaser}
\end{teaserfigure}

\begin{abstract}
In many recent works, multi-layer perceptions~(MLPs) have been shown to be suitable for modeling complex spatially-varying functions including images and 3D scenes.
\revised{Although the MLPs are able to represent complex scenes with unprecedented quality and memory footprint,} this expressive power of the MLPs, however, comes at the cost of long training and inference times. 
On the other hand, bilinear/trilinear interpolation on regular grid-based representations can give fast training and inference times, but \revised{cannot match the quality of MLPs without requiring significant additional memory.}
Hence, in this work, we investigate what is the \textit{smallest} change to grid-based representations that allows for retaining the high fidelity result of MLPs while enabling fast reconstruction and rendering times. 
We introduce a surprisingly simple change that achieves this task -- \textit{simply allowing a fixed non-linearity (ReLU) on interpolated grid values}. 
When combined with coarse-to-fine optimization, we show that such an approach becomes competitive with the state-of-the-art.
We report results on radiance fields, and occupancy fields, and compare against multiple existing alternatives. 
\revised{Code and data for the paper are available at \url{https://geometry.cs.ucl.ac.uk/projects/2022/relu_fields}.}
\end{abstract}

\maketitle

\mysection{Introduction}{Introduction}

Coordinate-based \acp{MLP} have been shown to be capable of representing complex signals with high fidelity and a low memory footprint. 
Exemplar applications include NeRF~\cite{mildenhall2020nerf}, which encodes  lighting-baked volumetric radiance-density field into a single \ac{MLP} using posed images; LIFF~\cite{LIFF2020}, which encodes 2D image signal into a single \ac{MLP} using multi-resolution pixel data. Alternatively, a 3D shape can be encoded as
\revised{
an occupancy field~\cite{mescheder2019occupancy,chen2019learning} or as a signed distance field~\cite{park2019deepsdf}.
}

A significant drawback of such approaches is that \acp{MLP} are both slow to train and slow to evaluate, especially for applications that require multiple evaluations per signal-sample (\eg multiple per-pixel evaluations during volume tracing in NeRFs).
On the other hand, traditional data structures like $n$-dimensional grids are fast to optimize and evaluate, but require a significant amount of memory to represent high frequency content
(see Figure~\ref{fig:quantitative_synth}).  
As a result, there has been an explosion of interest in hybrid representations that combine fast-to-evaluate data structures with coordinate-based MLPs, e.g., by encoding latent features in regular~\cite{sun2021direct} and adaptive~\cite{liu2020neural, aliev2020neural, martel2021acorn, mueller2022instant} grids and decoding linearly interpolated ``neural" features with a small MLP. 

In this paper, we revisit regular grid-based models and look for the \textit{minimum} change needed to make such grids perform on par with ``neural'' representations.
As the key takeaway message, we find that simply using a \ac{ReLU} non-linearity on top of interpolated grid values, without any additional learned parameters, optimized in a progressive manner already does a surprisingly good job, with minimal added complexity. For example, in Figure~\ref{fig:Teaser} we show results in the context of representing volumes (left) and surfaces (right) and on regularly sampled grid vertices for reconstruction, respectively. 
As additional benefits, these grid based 3D-models are amenable to generative modeling, and to local manipulation. 

In summary, we present the following contributions: 
(i)~we propose a minimal extension to grid-based signal representations, which we refer to as \names; 
(ii)~we show that this representation is \textit{simple}, does \textit{not} require any neural networks, is directly \textit{differentiable} (and hence easy to optimize), 
and is fast to  \textit{optimize and evaluate} (i.e. render); and (iii)~we empirically validate our claims by showing  applications where \names plug in naturally: first, image-based 3D scene reconstruction; and second, implicit modeling of 3D geometries.

\mysection{Related Work}{Related Work}

\paragraph{Discrete sample based representations}
Computer vision and graphics have long experimented with different representations for \revised{working with} visual data. \revised{While working with,} images are ubiquitously \revised{represented as} 2D grids of pixels, while due to the memory requirements; 3D models are often \revised{represented (and stored)} in a sparse format, e.g., as meshes, or as point clouds.
In the context of images, since as early as the sixties~\cite{billingsley1966processing}, different ideas have been proposed to make pixels more expressive.
One popular option is to store a fixed number (\eg one) of zero-crossing for explicit edge boundary information \cite{bala2003combining,tumblin2004bixels,ramanarayanan2004feature,laine2010efficient}, by using curves  \cite{parilov2008real}, or augmenting  pixels/voxels with more than one color~\cite{pavic2010two,agus2010split}. 
Another idea is to deform the underlying pixel grid by explicitly storing discontinuity information along general curves \cite{tarini2005pinchmaps}.
\citet{loviscach2005efficient} optimized MIP maps, such that the thresholded values match a reference.
Similar ideas were also being explored for textures and shadow maps~\cite{sen2003shadow,sen2004silhouette}, addressing specific challenges in sampling.
\name grid implicitly stores discontinuity information by varying grid values such that when interpolated and passed through a ReLU it represents a zero crossing per grid cell.

\myfigure{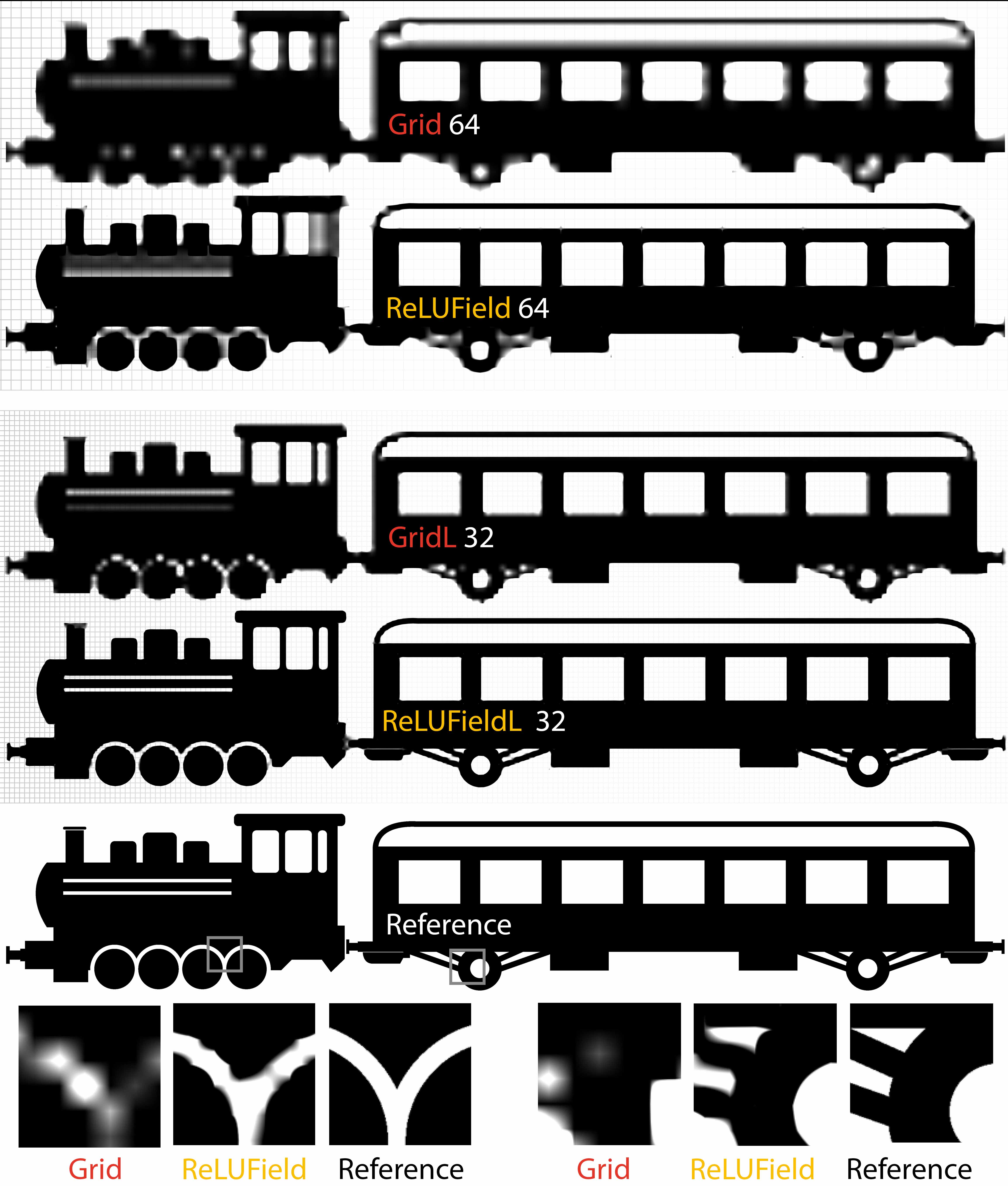}{
Representing an image with a standard pixel grid bi-linearly interpolated to a larger size (\method{Grid}) versus a \name of the same size (\method{ReLUField}).
The grid-size of the variants, \method{ReLUField} and \method{Grid}, is 64x smaller; while of, \method{ReLUFieldL} and \method{GridL}, is 32x smaller than the source image-resolution \textit{along each dimension}. Note that the `L' variants have a bigger grid-size and hence less smaller than the GT raster image. Simply adding a ReLU allows for significantly more sharpness and detail to be expressed. Hence, we can say that the humble ReLU is truly \emph{the little non-linearity that could}.
}

In the 2D domain, the regular pixel grid format of images has proven to be amenable to machine learning algorithms because CNNs are able to naturally input and output regularly sampled 2D signals as pixel grids. As a result, these architectures can be easily extended to 3D to operate on voxel grids, and therefore can be trained for many learning-based tasks, e.g., using differentiable volume rendering as supervision~\cite{tulsiani2017multi,henzler2019escaping,nguyen2019hologan,sitzmann2019deepvoxels}. However, such methods are inefficient with respect to memory and are hence typically restricted to low spatial resolution.

\paragraph{Learned neural representations}
Recently, coordinate-based MLPs representing continuous signals have been shown to be able to dramatically increase the representation quality of 3D objects~\cite{atlasNet:2018} or reconstruction quality of 3D scenes~\cite{mescheder2019occupancy,mildenhall2020nerf}. 
However, such methods incur a high computational cost, as the \ac{MLP} has to be evaluated, often multiple times, for each output signal location (\eg pixel) when performing differentiable volume rendering~\cite{mildenhall2020nerf,chan2021pi,schwarz2020graf,niemeyer2021giraffe}.
In addition, this representation is not well suited for post-training manipulations as the weights of the \ac{MLP} have a global effect on the structure of the scene.
To fix the slow execution, sometimes grid-like representations are fit post-hoc to a trained \ac{NeRF} model \cite{reiser2021kilonerf,hedman2021baking,yu2021plenoctrees,garbin2021fastnerf}, however such methods are unable to reconstruct scenes from scratch.

As a result, there has been an interest in hybrid methods that store learned features in spatial data structures, and accompany this with an \ac{MLP}, often much smaller, for decoding the interpolated neural feature signal at continuous locations.
Examples of such methods store learned features on regular grids~\cite{sitzmann2019deepvoxels,blockGAN:2020}, sparse voxels~\cite{liu2020neural,martel2021acorn}, point clouds~\cite{aliev2020neural}, local crops of 3D grids~\cite{jiang2020lig}, or on intersecting axis-aligned planes (triplane)~\cite{chan2021efficient}.

\paragraph{Concurrent work}
Investigating representations suitable for efficiently representing complex signals is an active area of research.
In this section, we discuss three concurrent works: \methodb{DVGo}~\cite{sun2021direct}, \methodb{Plenoxels}~\cite{yu2021plenoxels} and \methodb{NGP}~\cite{mueller2022instant}.

Reporting a finding similar to ours, \revised{\methodb{DVGo} proposes the use of a ``post-activated" (i.e.,  after interpolation) density grid for modelling high-frequency geometries. They model the view-dependent appearance through a learned feature grid which is decoded using an MLP. They in-fact show comprehensive experimental evaluation, on multiple datasets comparing to multiple baselines, for the task of image-based 3D scene reconstruction.}

\revised{\methodb{Plenoxels} proposes the use of sparse grid structure for modeling the scene with ReLU activation and, similar to our experiments, also uses spherical harmonic coefficients \cite{yu2021plenoctrees} for modeling  view-dependent appearance. }

\revised{\methodb{NGP}~\cite{mueller2022instant} proposes a hierarchical voxel-hashing scheme to store learned features and using a small MLP decoder for converting them into geometry and appearance. Their reconstruction-times are about significantly lower than the others  because of their impressively engineered GPU (low-level cuda) implementation.}

\revised{We believe that} our work differs from these concurrent efforts in that, our motivation is to investigate the \emph{minimal} change to existing voxel grids that \revised{can boost the per-capita signal modelling capacity of the grids when the signals contain sharp c1-discontinuities}. And hence as such, we are not focused only on 3D scene reconstruction, and similar to \methodb{NGP}, also consider other applications where grids are the de-facto representation, where \names might help. Our method is orthogonal to, and fully compatible with, the sparse data structures proposed in \methodb{Plenoxels} and \methodb{NGP}, and we expect the improvements gained by such approaches to be directly applicable to our work. The power and complexity of other methods, however, comes at the cost of not being able to load the resulting assets into legacy 3D modelling or volume visualization software (backward-compatibility), which is possible for our results, as long as the software can load signed data and apply transfer functions.

\myfigure{Concept}{Representing a ground-truth function \textbf{(blue)} in a 1D (a) and 2D (b) grid cell using the linear basis \textbf{(yellow)} and a \names \textbf{(pink)}.
The reference has a \revised{c1}-discontinuity inside the domain that a linear basis cannot capture.
A \name will pick two values $y_1$ and $y_2$, such that their interpolation, after clamping will match the sharp \revised{c1}-discontinuity in the ground-truth \textbf{(blue)} function.
}

\mysection{Method}{Method}
\subsection*{It's Just a Little ReLU}

\mymath{\function}f
\mymath{\signalDimension}{n}
\mymath{\domainDimension}{m}
\mymath{\gridSize}{r}
\mymath{\location}{\mathbf x}
\mymath{\grid}{G}
\mymath{\direction }{\omega}

We look for a representation of \signalDimension-valued signals 
on an \domainDimension-dimensional coordinate domain $\mathbb R^\domainDimension$. \revised{For simplicity, we explain the method for $\domainDimension=3$.} Our representation is strikingly simple. We consider a regular ($\domainDimension=3$)-dimensional ($\gridSize\times\gridSize\times\gridSize$)-grid \grid composed of \gridSize voxels along each side. Each voxel has a certain size defined by its diagonal norm in the ($\domainDimension=3$)-dimensional space and holds an \signalDimension-dimensional vector at each  of its ($2^{\domainDimension=3}=8$) vertices. Importantly, \revised{even though they have matching number of dimensions}, these values do not have a direct physical interpretation (\eg color, density, or occupancy), which always have some \revised{explicitly-defined} range, \eg $[0,1]$ or $\left[0,+\infty\right)$.
Rather, we store unbounded values on the grid; and thus for technical correctness, we call these grids ``feature''-grids instead of signal-grids.
The features at grid vertices are then interpolated using ($m=3$)-linear interpolation, and followed by a \textit{single non-linearity}: the \ac{ReLU}, \ie function $\mathtt{ReLU}(x) = \max(0, x)$ which maps negative input values to 0 and all other values to themselves. Note that this approach does not have any MLP or other neural-network that interprets the features, instead they are simply clipped before rendering. 
Intuitively, during optimization, these feature-values at the vertices can go up or down such that the ReLU clipping plane best aligns with the \revised{c1-}discontinuities within the ground-truth signal. Figure~\ref{fig:Concept} illustrates this concept.

\begin{algorithm}[b!]
    \caption{Fetching a 2D ReLU field.}
    \label{alg:Main}
    \begin{algorithmic}[1]
        \Procedure{ReluField2D}{$\grid, \location$}
            \State 
            $\location_\mathrm g$
            := 
            \Call{Floor}{\location}
            \State 
            $\location_\mathrm f$
            := 
            \Call{Frac}{\location}
            \State $y_{00}$ := 
            \Call{Fetch}
            {\grid, $\location_\mathrm g$ + (0,0)}
            \State $y_{01}$ := 
            \Call{Fetch}
            {\grid, $\location_\mathrm g$ + (0,1)}
            \State $y_{10}$ := 
            \Call{Fetch}
            {\grid, $\location_\mathrm g$ + (1,0)}
            \State $y_{11}$ := 
            \Call{Fetch}
            {\grid, $\location_\mathrm g$ + (1,1)}
            \State 
            y
            :=
            \Call
            {BiLinear}
            {$y_{00}$, $y_{01}$, $y_{10}$, $y_{11}$, $\location_\mathrm f$}
            \State\Return \Call{relu}{y}
        \EndProcedure
    \end{algorithmic}
\end{algorithm}

As a didactic example, we fit an image into a 2D \name grid similar to \cite{sitzmann2020implicit}, where grid values are stored as floats in the $(-\infty,+\infty)$ range. 
For any query position, we interpolate the grid values before passing through the ReLU function (see Algorithm~\ref{alg:Main}).
Since the image-signal values are expected to be in the $[0, 1]$ range, we apply a hard-upper-clip on the interpolated values just after applying the ReLU. 
\revised{
We can see in \refFig{Image} that \name allows us to represent sharp edges at a higher fidelity than bilinear interpolation (without the ReLU) at the same resolution grid size.
One limitation of this representation is that it can only well represent signals that have sparse c1-discontinuities, such as this flat-shaded images and as we show later, 3D volumetric density.
However, other types of signals, such as natural images, do not benefit from using a \names representation (see supplementary material).
}

\begin{figure*}[t!]
    \centering
    \includegraphics[width=.98\textwidth]{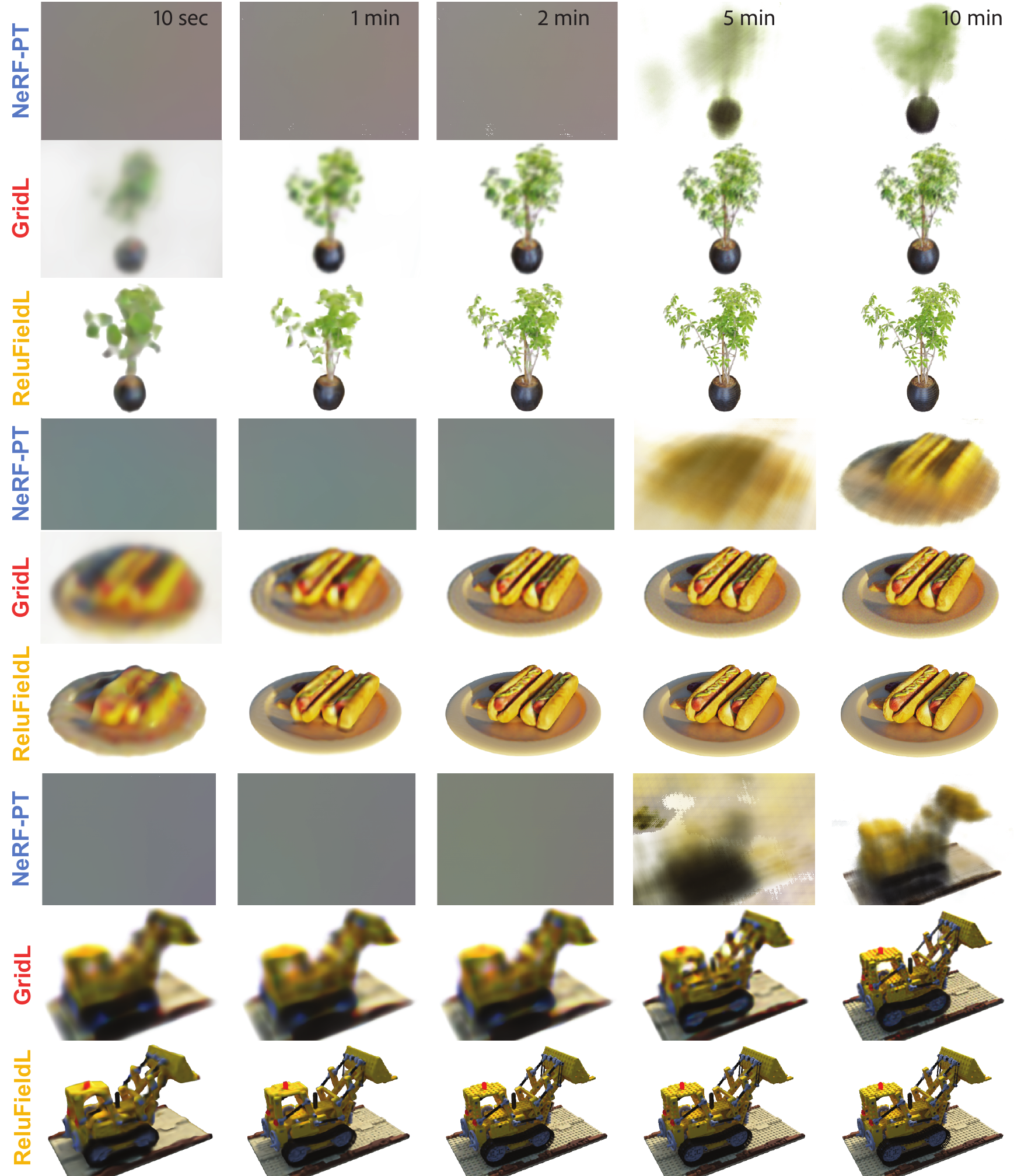}
    \caption{Qualitative comparison between \method{NeRF-PT}, \method{GridL} and \method{ReLUFieldL}. Grid-based versions converge much faster, and we can see significant sharpness improvements of \method{ReLUFieldL} over \method{GridL}, for example in the leaves of the plant. See also supplementary video. }
    \label{fig:quantitative_synth}
\end{figure*}

\mysection{Applications}{Applications}
We now demonstrate two different applications of \names; NeRF-like 3D scene-reconstruction (\refSec{Radiance}), and 3D object reconstruction via occupancy fields (\refSec{Occupancy}).

\mysubsection{Radiance Fields}{Radiance}

\mymath{\inputImage}{I}
\mymath{\renderedImage}{\hat{I}}
\mymath{\numberOfInputImages}{n}
\mymath{\inputImages}{\mathcal I}
\mymath{\poses}{\mathcal C}
\mymath{\pose}{\mathsf C}
\mymath{\representation}{\mathcal{S}}
\mymath{\rendering}{\mathcal R}
\mymath{\rotation}{R}
\mymath{\translation}{T}
\mymath{\height}{H}
\mymath{\width}{W}
\mymath{\focal}{F}
\mymath{\realNumberSet}{\mathbb{R}}
\mymath{\reluFunction}{\mathtt{relu}}
\mymath{\gridOpt}{\grid^{*}}

In this application, we discuss how \name can be used in place of the coordinate-based MLP in NeRF~\cite{mildenhall2020nerf}.
Input to this approach are a set of images $\inputImages=\{\inputImage_1, \ldots, \inputImage_\numberOfInputImages \}$ and corresponding camera poses $\poses=\{\pose_1,  \ldots, \pose_\numberOfInputImages\}$, where each camera pose consists of $\pose=\{\rotation, \translation, \height, \width, \focal\}$; \rotation is the rotation-matrix ($\rotation \in \realNumberSet^{ 3 \times 3}$), \translation is the translation-vector ($\translation \in \realNumberSet^3$), \height, \width are the scalars representing the height and width respectively, and \focal denotes the focal length. We assume that the respective poses for the images are known either through hardware calibration or by using structure-from-motion~\cite{schonberger2016structure}.

We denote the rendering operation to convert the 3D scene representation \representation and the camera pose \pose into an image as $\rendering(\representation, \pose)$.
Thus, given the input set of images \inputImages and their corresponding camera poses \poses, the problem is to recover the underlying 3D scene representation \representation such that when rendered from any $\pose_i \in \poses$, \representation produces rendered image $\renderedImage_i$ as close as possible to the input image $\inputImage_i$, and produces spatio-temporally consistent $\renderedImage_j$ for poses $\pose_j \not\in \poses$.

\paragraph{Scene representation}
We model the underlying 3D scene representation \representation, which is to be recovered, by a \name.
The vertices of the grid store, first, raw pre-\reluFunction density values in $(-\infty,\infty)$ that model geometry, and, second, the second-degree \ac{SH} coefficients\revised{~\cite{yu2021plenoctrees, wizadwongsa2021nex}} that model view-dependent appearance.
The \reluFunction is only applied to pre-\reluFunction density, not to appearance.

We directly optimize values at the vertices to minimize the photometric loss between the rendered images \renderedImage and the input images \inputImage.
The optimized grid \gridOpt, corresponding to the recovered 3D scene \representation, is obtained as:
\vspace{-.2cm}
\begin{equation}
\gridOpt  = 
\operatorname{arg\,min}_\grid
\sum_{i=1}^\numberOfInputImages
\| \inputImage_i -
\overbrace{ 
\rendering(\grid, \pose_i)
}^\text{\footnotesize {$\hat{I}_i$}}
\|^2_2. 
\end{equation}

\paragraph{Implementation details}
\label{sec:Implementation_details}
Similar to \methodb{NeRF}, we use the EA (emission-absorption) raymarching model~\cite{max1995optical, henzler2019escaping, mildenhall2020nerf} for realizing the rendering function \rendering.
The grid is scaled to a single global AABB (Axis-Aligned-Bounding-Box) that is encompassed by the camera frustums of all the available poses \poses, and is initialized with uniform random values. 
We optimize the vertex values using Adam \cite{kingma2014adam} with a learning rate of 0.03, and all other default values, for all examples shown.

We perform the optimization progressively in a coarse-to-fine manner similar to \citet{karras2018progressive}. 
Initially, the feature grid is optimized at a resolution where each dimension is reduced by a factor of $2^4$.
After a fixed number of iterations at each stage $N$, the grid resolution is doubled and the features on the feature-grid \grid are tri-linearly upsampled to initialize the next stage.
This proceeds until the final target resolution is reached.

\newcommand{\winner}[1]{\textbf{#1}}
\newcommand{\runnerup}[1]{\underline{#1}}
\begin{table*}[]
    \centering
    \setlength{\tabcolsep}{0.1cm}
    \captionof{table}{Evaluation results on 3D synthetic scenes. Metrics used are PSNR ($\uparrow$) / LPIPS ($\downarrow$). \revised{The column \method{NeRF-TF}$^*$ quotes PSNR values from prior work~\cite{mildenhall2020nerf}, and as such we do not have a comparable runtime for this method.}}
    \begin{tabular}{r rr rr rr rr rr rr rr rr rr}
         Scene &
         \multicolumn2c{\method{NeRF-TF}$^*$} &
         \multicolumn2c{\method{NeRF-PT}} &
         \multicolumn2c{\method{Grid}} &
         \multicolumn2c{\method{GridL}} &
         \multicolumn2c{\method{ReLUField}} &
         \multicolumn2c{\method{ReLUFieldL}} &
         \multicolumn2c{\method{RFLong}} &
         \multicolumn2c{\method{RFNoPro}} \\
         \cmidrule(lr){2-3}
         \cmidrule(lr){4-5}
         \cmidrule(lr){6-7}
         \cmidrule(lr){8-9}
         \cmidrule(lr){10-11}
         \cmidrule(lr){12-13}
         \cmidrule(lr){14-15}
         \cmidrule(lr){16-17}
         &
         \multicolumn1c{\footnotesize{PSNR}}&
         \multicolumn1c{\footnotesize{LPIPS}}&
         \multicolumn1c{\footnotesize{PSNR}}&
         \multicolumn1c{\footnotesize{LPIPS}}&
         \multicolumn1c{\footnotesize{PSNR}}&
         \multicolumn1c{\footnotesize{LPIPS}}&
         \multicolumn1c{\footnotesize{PSNR}}&
         \multicolumn1c{\footnotesize{LPIPS}}&
         \multicolumn1c{\footnotesize{PSNR}}&
         \multicolumn1c{\footnotesize{LPIPS}}&
         \multicolumn1c{\footnotesize{PSNR}}&
         \multicolumn1c{\footnotesize{LPIPS}}&
         \multicolumn1c{\footnotesize{PSNR}}&
         \multicolumn1c{\footnotesize{LPIPS}}&
         \multicolumn1c{\footnotesize{PSNR}}&
         \multicolumn1c{\footnotesize{LPIPS}}
         \\
         \toprule
         \scene{Chair}
         & \runnerup{33.00} & \runnerup{0.04} &
         \winner{33.75} & 
         \winner{0.03} & 
         25.53 & 
         0.12 & 
         27.08 & 
         0.11 & 
         31.50 & 
         0.05 & 
         32.39 & 
         0.03 & 31.77 & 0.05  & 13.85 & 0.48 \\
         \scene{Drums} & \runnerup{25.01} & \runnerup{0.09} & 23.82 & 0.12 & 19.85 & 0.20 & 20.70 & 0.17 & 23.13 & 0.09 & \winner{25.15} & \winner{0.06} & 23.78 & 0.09  & 10.74 & 0.52 \\
         \scene{Ficus} & \winner{30.13} & \winner{0.04} & \runnerup{28.96} & 0.04 & 22.10 & 0.13 & 23.61 & 0.11 & 25.89 & 0.06 & 27.37 & \runnerup{0.04} & 26.11 & 0.05  & 13.21 & 0.47\\
         \scene{Hotdog} & \winner{36.18} & 0.12 & 33.52 & \runnerup{0.06} & 28.53 & 0.12 & 29.83 & 0.10 & 34.65 & 0.03 & \runnerup{35.72} & \winner{0.03} & 34.70 & 0.03  & 12.22 & 0.53\\
         \scene{Lego} & \winner{32.54} & \runnerup{0.05} & 28.36 & 0.08 & 23.76 & 0.17 & 23.97 & 0.15 & 28.83 & 0.06 & \runnerup{30.78} & \winner{0.03} & 29.64 & 0.05  & 10.63 & 0.56\\
         \scene{Materials} & \winner{29.62} & 0.06 & \runnerup{29.23} & \winner{0.04} & 21.87 & 0.18 & 22.74 & 0.13 & 27.41 & 0.06 & 28.23 & \runnerup{0.05} & 28.23 & 0.05  &  8.99 & 0.55  \\
         \scene{Mic} & \runnerup{32.91} & 0.02 & \winner{33.08} & 0.02 & 25.87 & 0.08 & 25.91 & 0.08 & 31.88 & 0.03 & 32.62 & 0.02 & 31.22 & 0.03  & 12.47 & 0.41  \\
         \scene{Ship} & \runnerup{28.65} & 0.20 & \winner{29.22} & 0.14 & 23.86 & 0.25 & 22.54 & 0.24 & 26.86 & 0.14 & 28.02 & 0.12 & 27.39 & 0.13  &  9.92 & 0.59 \\
         \midrule
         Average & \winner{31.01} & 0.07 & 29.99 & 0.07 & 23.92 & 0.16 & 24.54 & 0.14 & 28.77 & 0.07 & \runnerup{30.04} & \winner{0.05} & 29.10 & \runnerup{0.06} & 11.50 & 0.51 \\
         \midrule
         Time (recon)&
         \multicolumn2c{---} &
         \multicolumn2c{11h:21m:00s} & 
         \multicolumn2c{00h:03m:41s} & 
         \multicolumn2c{00h:10m:02s} & 
         \multicolumn2c{00h:03m:41s} & 
         \multicolumn2c{00h:10m:36s} &
         \multicolumn2c{10h:51m:29s} &
         \multicolumn2c{00h:07m:11s}
         \\
         Time (render)&
         \multicolumn2c{---} &
         \multicolumn2c{16,363.0\,ms}&
         \multicolumn2c{9.0\,ms}&
         \multicolumn2c{99.1\,ms}&
         \multicolumn2c{9.1\,ms}&
         \multicolumn2c{99.5\,ms}&
         \multicolumn2c{9.1\,ms} &
         \multicolumn2c{9.1\,ms}
         \\
         \bottomrule
    \end{tabular}
    \label{tab:QuantitativeSynthetic}
\end{table*}

\paragraph{Evaluation}
We perform experiments on the eight synthetic Blender scenes used by NeRF \cite{mildenhall2020nerf}, viz. \scene{Chair}, \scene{Drums}, \scene{Ficus}, \scene{Hotdog}, \scene{Lego}, \scene{Materials}, \scene{Mic}, and \scene{Ship} and compare our method to prior works, baselines, and ablations. We also show an extension of \names to one of their real world captured scenes, named \scene{Flowers}.

First, we compare to the mlp-based baseline NeRF \cite{mildenhall2020nerf}. For the purpose of these experiments though, we use the public \texttt{nerf-pytorch} version~\cite{nerfpytorch} for comparable training-time comparisons since all our implementations are in PyTorch. \revised{For disambiguation, we refer to this PyTorch version as \method{NeRF-PT} and the original one as \method{NeRF-TF} and report scores for both.}
Second, we compare to two versions of traditional grids where vertices \revised{store scalar density and second-degree \ac{SH} approximations of the appearance}, namely \method{Grid} (\ie $128^3$ grid) and \method{GridL} (\ie $256^3$ grid).
Finally, we compare to our approach at the same two resolutions, \method{ReLUField} and \method{ReLUFieldL}. 
The above four methods are optimized with the same progressive growing setting with $N=2000$, and all the same hyperparameters except the grid resolution.
We report PSNR and LPIPS \cite{zhang2018unreasonable} computed on a held-out test-set of Image-Pose pairs different from the training-set (\inputImages, \poses).
All training times were recorded on 32GB-V100 GPU while the inference times were computed on RTX~2070 Super. 
Our method is implemented entirely in PyTorch and does not make use of any custom GPU kernels.

Table \ref{tab:QuantitativeSynthetic} summarizes results from these experiments.
We can see that traditional physically-based grid baselines \method{Grid} and \method{GridL} perform the worst, while our method has comparable performance to \method{NeRF-PT} and is much faster to reconstruct and render.
This retains the utility of grid-based models for real-time applications without compromising on quality. Figure \ref{fig:quantitative_synth} demonstrates qualitative results from these experiments.

\paragraph{Ablations} 
We ablate the components described in \ref{sec:Implementation_details}, and and also include the results in Table~\ref{tab:QuantitativeSynthetic} in the last two columns. 
\method{RFLong} is a normal \name optmized for a much longer time (comparable to \method{NeRF-PT}'s training time).
We see minor improvement over the default settings, however we can see that the optimization time plays less of a role than the resolution of the grid itself (\method{ReLUFieldL} outperforms \method{RFLong}).
\method{RFNoPro} is trained without progressive growing for the same number of total steps. 
We see that it yields a much lower reconstruction quality, indicating that progressive growing is critical for the grid to converge to a good reconstruction.

\begin{figure}[h!]
   \includegraphics[width=\columnwidth]{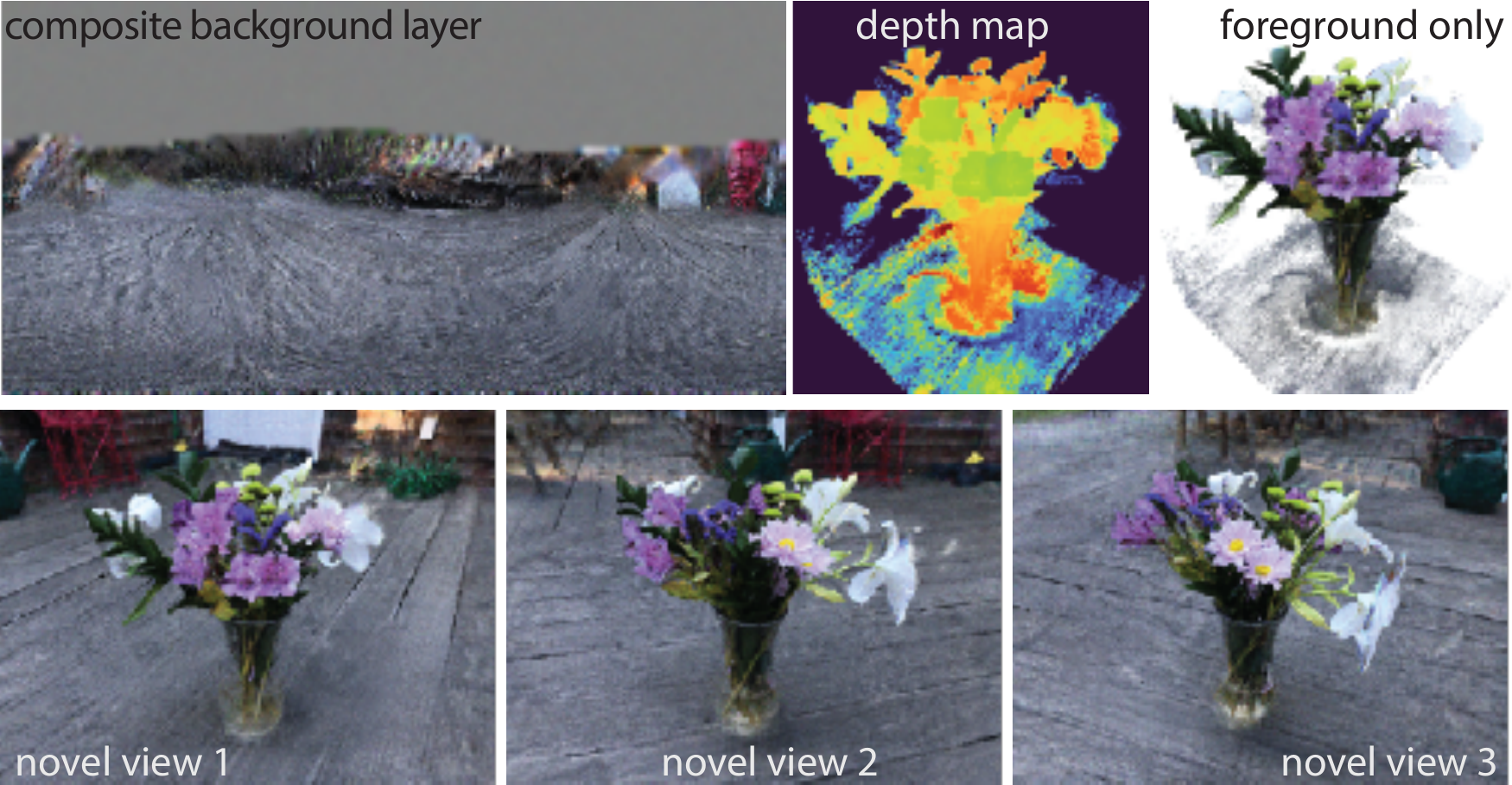}
   \caption{Qualitative results for the real-captured scene extension of \names on \scene{Flowers}. We decompose the scene into a series of spherical-background shells and a foreground \name layer, which are alpha-composited together to give final novel view renderings. The top-left visualization shows the composite of the background spherical shells un-projected onto a 2D image-plane.
   }
   \label{fig:QualitativeReal}
\end{figure}

\paragraph{Real scene extension}
Similar to the real-captured-360 scenes from the \methodb{NeRF}, we also show an extension of \names to modeling real scenes.
In this example, we model the background using a ``MultiSphereGrid'' representation, as proposed by \citet{attal2020matryodshka}. \revised{Please note that the background grid is modeled as a regular bilinear grid without any  ReLU.} For simplicity, we use an Equi-rectangular projection~(ERP) instead of Omni-directional stereo~(ODS) for mapping the Image-plane to the set of background spherical shells. 
Fig. \ref{fig:QualitativeReal} shows qualitative results for this extensions after one hour of optimization.
Here, we can see that the grid does a good job of representing the complex details in the flower, while the background is modeled reasonably well by the shells.

\mysubsection{Occupancy Fields}{Occupancy}
Another application of coordinate-based \acp{MLP} is as a representation of (watertight) 3D geometry. Here, we fit a high resolution ground-truth mesh, as a 3D occupancy field~\cite{mescheder2019occupancy} into a \name.
One might want to do this in order to, for example, take advantage of the volumetric-grid structure to learn priors over geometry, something that is harder to do with meshes or coordinate-based \acp{MLP} directly.

\paragraph{Occupancy representation}
The core \name representation used for this application only differs from the radiance fields setup (see \refSec{Radiance}) as follows: 
First, since we are only interested in geometry, we do not store any \ac{SH} coefficients on the grid, and simply model volumetric occupancy as a probability from $[0,1]$.
Second, as supervision, we use ground truth point-wise occupancy values in 3D (\ie 1, if the point lies inside the mesh, and 0 otherwise), rather than rendering an image and applying the loss on the rendered image. 
Finally, since the ground truth occupancy values are binary, we use a binary cross entropy~(BCE) loss.
Thus, we obtain the optimized grid $G^*$ as,
\begin{equation}
G^*  :=  
\operatorname{arg\,min}_\grid
\sum_{\mathbf{x} \in \mathcal{B}}
\text{BCE}(O(\mathbf{x}),  \textsc{ReluField3D} (\tanh(G), \mathbf{x}) )
\end{equation}
where, $O$ is the ground truth occupancy,
$\mathbf{x}$ denote sample locations inside an axis-aligned bounding box $\mathcal{B}$, 
BCE denotes the binary cross entropy loss, 
and $G$ represents the \name grid.
Note that we use the $\tanh$ to limit the grid values in $(-1,1)$, although other bounding functions, or tensor-normalizations can be used.

We initialize the grid with uniform random values.
The supervision signal comes from sampling random points inside and around the tight AABB of the GT high resolution mesh, and generating the occupancy values for those points by doing an inside-outside test on the fly during training.
For rendering, we directly show the depth rendering of the obtained occupancy values.
We define the grid-extent and the voxel size by obtaining the \textit{AABB} ensuring a tight fit around the GT mesh. 
\revised{
\begin{table}[h]
    \setlength{\tabcolsep}{10pt}
    \centering
    \caption{\revised{Evaluation results on modeling 3D geometries as occupancy fields. Metric used is Volumetric-IoU \cite{mescheder2019occupancy}. The baseline \method{MLP} is  our implementation of OccupancyNetworks \cite{mescheder2019occupancy}}.}
    \label{tab:Occupancy}
    \begin{tabular}{rrrr}
            &
            \multicolumn1c{\method{MLP}}&
            \multicolumn1c{\method{Grid}}&
            \multicolumn1c{\method{ReLUField}} \\
            \toprule
            Thai Statue & \runnerup{0.867} & 0.827 & \winner{0.901} \\
            Lucy & \runnerup{0.920} & 0.883 & \winner{0.935} \\
            Bimba & \runnerup{0.983} & 0.978 & \winner{0.987} \\
            Grog & \runnerup{0.961} & 0.947 & \winner{0.971} \\ 
            Lion & 0.956 & \runnerup{0.970} & \winner{0.979} \\
            Ramses & \runnerup{0.973} & 0.961 & \winner{0.978} \\
            Dragon & \runnerup{0.886} & 0.761 & \winner{0.896} \\
            \toprule
            Average volumetric-IoU & \runnerup{0.935} & 0.903 & \winner{0.949} \\ 
        \bottomrule
    \end{tabular}
\end{table}
}

\mycfigure{QualitativeOccupancy_newer}
{Qualitative results for the occupancy fields  comparing \method{Grid}, \method{MLP}, and \method{ReLUField}.}

\paragraph{Evaluation}
Figure~\ref{fig:QualitativeOccupancy_newer} shows the qualitative results of the different representations used for this task. We can see that a \method{ReLUField} in this case yields higher quality reconstructions than a standard \method{Grid}, or a coordinate-based \method{MLP}.
\revised{
Quantitative scores, Volumetric-IoU as used in \cite{mescheder2019occupancy}, for the \scene{ThaiStatue}, \scene{Lucy}, \scene{Bimba}, \scene{Grog}, \scene{Lion}, \scene{Ramses}, and \scene{Dragon} models are summarized in \refTab{Occupancy}.
\method{ReLUField} and \method{Grid} require 15 mins, while \method{MLP} requires 1.5~hours for training.
}

\mysection{Discussion}{Discussion}
\mysubsection{Limitations}{Limitations}
Our approach has some limitations. 
First, the resulting representations are large. 
A \name of size $128^3$ used for radiance fields (\ie with SH coefficients) takes 260Mb, and the large version at $256^3$ takes 2.0\,Gb of storage. 
We believe that combining \name with a sparse data structure would see significant gains in performance and reduction in the memory footprint. 
However, in this work we emphasize the simplicity of our approach and show that the single non-linearity alone is responsible for a surprising degree of quality improvement.

\name also cannot model more than one ``crease'' (i.e., discontinuity) per grid cell. While learned features allow for more complex signals to be represented, they do so at the expense of high compute costs. 
The purpose of this work is to refocus attention on \textit{what is actually required for high fidelity scene reconstruction}. 
We believe that the task definition and data are responsible for the high quality results we are seeing now, and show that traditional approaches can yield good results with minor modifications, and neural networks may not be required.
However, this is just one data-point in the space of possible representations, for a given specific task we expect that the optimal representation may be a combination of learned features, neural networks, and discrete signal representations. 

\mysubsection{Conclusion}{Conclusion}
In summary, we presented \name, an almost embarrassingly simple approach for representing signals; \textit{storing unbounded data on N-dimensional grid, and applying a single ReLU after linear interpolation}. 
This change can be incorporated at virtually no computational cost or complexity on top of existing grid-based methods, and strictly improve their representational capability. 
Our approach contains only values at grid vertices which can be directly optimized via gradient descent; does not rely on any  learned parameters, special initialization, or neural networks; and performs comparably with state-of-the-art approaches in only a fraction of the time.

\begin{acks}
The authors would like to thank the reviewers for their valuable suggestions. The research was partially supported by the European Union’s Horizon 2020 research and innovation programme under the Marie Skłodowska-Curie grant agreement No. 956585, gifts from Adobe, and the UCL AI Centre.
\end{acks}

\newpage
\bibliographystyle{ACM-Reference-Format}
\bibliography{paper}


\begin{thebibliography}{47}


\ifx \showCODEN    \undefined \def \showCODEN     #1{\unskip}     \fi
\ifx \showDOI      \undefined \def \showDOI       #1{#1}\fi
\ifx \showISBNx    \undefined \def \showISBNx     #1{\unskip}     \fi
\ifx \showISBNxiii \undefined \def \showISBNxiii  #1{\unskip}     \fi
\ifx \showISSN     \undefined \def \showISSN      #1{\unskip}     \fi
\ifx \showLCCN     \undefined \def \showLCCN      #1{\unskip}     \fi
\ifx \shownote     \undefined \def \shownote      #1{#1}          \fi
\ifx \showarticletitle \undefined \def \showarticletitle #1{#1}   \fi
\ifx \showURL      \undefined \def \showURL       {\relax}        \fi
\providecommand\bibfield[2]{#2}
\providecommand\bibinfo[2]{#2}
\providecommand\natexlab[1]{#1}
\providecommand\showeprint[2][]{arXiv:#2}

\bibitem[\protect\citeauthoryear{??}{ner}{2021}]%
        {nerfpytorch}
 \bibinfo{year}{2021}\natexlab{}.
\newblock \bibinfo{title}{Nerf-Pytorch}.
\newblock
  \bibinfo{howpublished}{\url{https://github.com/yenchenlin/nerf-pytorch}}.
\newblock


\bibitem[\protect\citeauthoryear{Agus, Gobbetti, Guiti{\'a}n, and Marton}{Agus
  et~al\mbox{.}}{2010}]%
        {agus2010split}
\bibfield{author}{\bibinfo{person}{Marco Agus}, \bibinfo{person}{Enrico
  Gobbetti}, \bibinfo{person}{Jos{\'e} Antonio~Iglesias Guiti{\'a}n}, {and}
  \bibinfo{person}{Fabio Marton}.} \bibinfo{year}{2010}\natexlab{}.
\newblock \showarticletitle{Split-Voxel: A Simple Discontinuity-Preserving
  Voxel Representation for Volume Rendering.}. In \bibinfo{booktitle}{\emph{VG@
  Eurographics}}. \bibinfo{pages}{21--28}.
\newblock


\bibitem[\protect\citeauthoryear{Aliev, Sevastopolsky, Kolos, Ulyanov, and
  Lempitsky}{Aliev et~al\mbox{.}}{2020}]%
        {aliev2020neural}
\bibfield{author}{\bibinfo{person}{Kara-Ali Aliev}, \bibinfo{person}{Artem
  Sevastopolsky}, \bibinfo{person}{Maria Kolos}, \bibinfo{person}{Dmitry
  Ulyanov}, {and} \bibinfo{person}{Victor Lempitsky}.}
  \bibinfo{year}{2020}\natexlab{}.
\newblock \bibinfo{title}{Neural Point-Based Graphics}.
\newblock
\newblock
\showeprint[arxiv]{cs.CV/1906.08240}


\bibitem[\protect\citeauthoryear{Attal, Ling, Gokaslan, Richardt, and
  Tompkin}{Attal et~al\mbox{.}}{2020}]%
        {attal2020matryodshka}
\bibfield{author}{\bibinfo{person}{Benjamin Attal}, \bibinfo{person}{Selena
  Ling}, \bibinfo{person}{Aaron Gokaslan}, \bibinfo{person}{Christian
  Richardt}, {and} \bibinfo{person}{James Tompkin}.}
  \bibinfo{year}{2020}\natexlab{}.
\newblock \showarticletitle{Matryodshka: Real-time 6dof video view synthesis
  using multi-sphere images}. In \bibinfo{booktitle}{\emph{European Conference
  on Computer Vision}}. Springer, \bibinfo{pages}{441--459}.
\newblock


\bibitem[\protect\citeauthoryear{Bala, Walter, and Greenberg}{Bala
  et~al\mbox{.}}{2003}]%
        {bala2003combining}
\bibfield{author}{\bibinfo{person}{Kavita Bala}, \bibinfo{person}{Bruce
  Walter}, {and} \bibinfo{person}{Donald~P Greenberg}.}
  \bibinfo{year}{2003}\natexlab{}.
\newblock \showarticletitle{Combining edges and points for interactive
  high-quality rendering}.
\newblock \bibinfo{journal}{\emph{ACM Transactions on Graphics (TOG)}}
  \bibinfo{volume}{22}, \bibinfo{number}{3} (\bibinfo{year}{2003}),
  \bibinfo{pages}{631--640}.
\newblock


\bibitem[\protect\citeauthoryear{Billingsley}{Billingsley}{1966}]%
        {billingsley1966processing}
\bibfield{author}{\bibinfo{person}{Fred~C Billingsley}.}
  \bibinfo{year}{1966}\natexlab{}.
\newblock \showarticletitle{Processing ranger and mariner photography}.
\newblock \bibinfo{journal}{\emph{Optical Engineering}} \bibinfo{volume}{4},
  \bibinfo{number}{4} (\bibinfo{year}{1966}), \bibinfo{pages}{404147}.
\newblock


\bibitem[\protect\citeauthoryear{Chan, Lin, Chan, Nagano, Pan, Mello, Gallo,
  Guibas, Tremblay, Khamis, Karras, and Wetzstein}{Chan et~al\mbox{.}}{2021a}]%
        {chan2021efficient}
\bibfield{author}{\bibinfo{person}{Eric~R. Chan}, \bibinfo{person}{Connor~Z.
  Lin}, \bibinfo{person}{Matthew~A. Chan}, \bibinfo{person}{Koki Nagano},
  \bibinfo{person}{Boxiao Pan}, \bibinfo{person}{Shalini~De Mello},
  \bibinfo{person}{Orazio Gallo}, \bibinfo{person}{Leonidas Guibas},
  \bibinfo{person}{Jonathan Tremblay}, \bibinfo{person}{Sameh Khamis},
  \bibinfo{person}{Tero Karras}, {and} \bibinfo{person}{Gordon Wetzstein}.}
  \bibinfo{year}{2021}\natexlab{a}.
\newblock \bibinfo{title}{Efficient Geometry-aware 3D Generative Adversarial
  Networks}.
\newblock
\newblock
\showeprint[arxiv]{cs.CV/2112.07945}


\bibitem[\protect\citeauthoryear{Chan, Monteiro, Kellnhofer, Wu, and
  Wetzstein}{Chan et~al\mbox{.}}{2021b}]%
        {chan2021pi}
\bibfield{author}{\bibinfo{person}{Eric~R Chan}, \bibinfo{person}{Marco
  Monteiro}, \bibinfo{person}{Petr Kellnhofer}, \bibinfo{person}{Jiajun Wu},
  {and} \bibinfo{person}{Gordon Wetzstein}.} \bibinfo{year}{2021}\natexlab{b}.
\newblock \showarticletitle{pi-gan: Periodic implicit generative adversarial
  networks for 3d-aware image synthesis}. In \bibinfo{booktitle}{\emph{IEEE
  CVPR}}. \bibinfo{pages}{5799--5809}.
\newblock


\bibitem[\protect\citeauthoryear{Chen, Liu, and Wang}{Chen
  et~al\mbox{.}}{2020}]%
        {LIFF2020}
\bibfield{author}{\bibinfo{person}{Yinbo Chen}, \bibinfo{person}{Sifei Liu},
  {and} \bibinfo{person}{Xiaolong Wang}.} \bibinfo{year}{2020}\natexlab{}.
\newblock \showarticletitle{Learning Continuous Image Representation with Local
  Implicit Image Function}.
\newblock \bibinfo{journal}{\emph{CoRR}}  \bibinfo{volume}{abs/2012.09161}
  (\bibinfo{year}{2020}).
\newblock


\bibitem[\protect\citeauthoryear{Chen and Zhang}{Chen and Zhang}{2019}]%
        {chen2019learning}
\bibfield{author}{\bibinfo{person}{Zhiqin Chen} {and} \bibinfo{person}{Hao
  Zhang}.} \bibinfo{year}{2019}\natexlab{}.
\newblock \showarticletitle{Learning implicit fields for generative shape
  modeling}. In \bibinfo{booktitle}{\emph{Proceedings of the IEEE/CVF
  Conference on Computer Vision and Pattern Recognition}}.
  \bibinfo{pages}{5939--5948}.
\newblock


\bibitem[\protect\citeauthoryear{Garbin, Kowalski, Johnson, Shotton, and
  Valentin}{Garbin et~al\mbox{.}}{2021}]%
        {garbin2021fastnerf}
\bibfield{author}{\bibinfo{person}{Stephan~J Garbin}, \bibinfo{person}{Marek
  Kowalski}, \bibinfo{person}{Matthew Johnson}, \bibinfo{person}{Jamie
  Shotton}, {and} \bibinfo{person}{Julien Valentin}.}
  \bibinfo{year}{2021}\natexlab{}.
\newblock \showarticletitle{Fastnerf: High-fidelity neural rendering at
  200fps}.
\newblock \bibinfo{journal}{\emph{arXiv preprint arXiv:2103.10380}}
  (\bibinfo{year}{2021}).
\newblock


\bibitem[\protect\citeauthoryear{Groueix, Fisher, Kim, Russell, and
  Aubry}{Groueix et~al\mbox{.}}{2018}]%
        {atlasNet:2018}
\bibfield{author}{\bibinfo{person}{Thibault Groueix}, \bibinfo{person}{Matthew
  Fisher}, \bibinfo{person}{Vladimir~G. Kim}, \bibinfo{person}{Bryan~C.
  Russell}, {and} \bibinfo{person}{Mathieu Aubry}.}
  \bibinfo{year}{2018}\natexlab{}.
\newblock \showarticletitle{AtlasNet: {A} Papier-M{\^{a}}ch{\'{e}} Approach to
  Learning 3D Surface Generation}.
\newblock \bibinfo{journal}{\emph{CoRR}}  \bibinfo{volume}{abs/1802.05384}
  (\bibinfo{year}{2018}).
\newblock


\bibitem[\protect\citeauthoryear{Hedman, Srinivasan, Mildenhall, Barron, and
  Debevec}{Hedman et~al\mbox{.}}{2021}]%
        {hedman2021baking}
\bibfield{author}{\bibinfo{person}{Peter Hedman}, \bibinfo{person}{Pratul~P
  Srinivasan}, \bibinfo{person}{Ben Mildenhall}, \bibinfo{person}{Jonathan~T
  Barron}, {and} \bibinfo{person}{Paul Debevec}.}
  \bibinfo{year}{2021}\natexlab{}.
\newblock \showarticletitle{Baking Neural Radiance Fields for Real-Time View
  Synthesis}.
\newblock \bibinfo{journal}{\emph{arXiv preprint arXiv:2103.14645}}
  (\bibinfo{year}{2021}).
\newblock


\bibitem[\protect\citeauthoryear{Henzler, Mitra, and Ritschel}{Henzler
  et~al\mbox{.}}{2019}]%
        {henzler2019escaping}
\bibfield{author}{\bibinfo{person}{Philipp Henzler}, \bibinfo{person}{Niloy~J
  Mitra}, {and} \bibinfo{person}{Tobias Ritschel}.}
  \bibinfo{year}{2019}\natexlab{}.
\newblock \showarticletitle{Escaping plato's cave: 3d shape from adversarial
  rendering}. In \bibinfo{booktitle}{\emph{ICCV}}. \bibinfo{pages}{9984--9993}.
\newblock


\bibitem[\protect\citeauthoryear{Jiang, Sud, Makadia, Huang, Nie{\ss}ner, and
  Funkhouser}{Jiang et~al\mbox{.}}{2020}]%
        {jiang2020lig}
\bibfield{author}{\bibinfo{person}{Chiyu Jiang}, \bibinfo{person}{Avneesh Sud},
  \bibinfo{person}{Ameesh Makadia}, \bibinfo{person}{Jingwei Huang},
  \bibinfo{person}{Matthias Nie{\ss}ner}, {and} \bibinfo{person}{Thomas
  Funkhouser}.} \bibinfo{year}{2020}\natexlab{}.
\newblock \showarticletitle{Local Implicit Grid Representations for 3D Scenes}.
  In \bibinfo{booktitle}{\emph{Proceedings of the IEEE Conference on Computer
  Vision and Pattern Recognition}}.
\newblock


\bibitem[\protect\citeauthoryear{Karras, Aila, Laine, and Lehtinen}{Karras
  et~al\mbox{.}}{2018}]%
        {karras2018progressive}
\bibfield{author}{\bibinfo{person}{Tero Karras}, \bibinfo{person}{Timo Aila},
  \bibinfo{person}{Samuli Laine}, {and} \bibinfo{person}{Jaakko Lehtinen}.}
  \bibinfo{year}{2018}\natexlab{}.
\newblock \showarticletitle{Progressive Growing of {GAN}s for Improved Quality,
  Stability, and Variation}. In \bibinfo{booktitle}{\emph{International
  Conference on Learning Representations}}.
\newblock
\urldef\tempurl%
\url{https://openreview.net/forum?id=Hk99zCeAb}
\showURL{%
\tempurl}


\bibitem[\protect\citeauthoryear{Kingma and Ba}{Kingma and Ba}{2014}]%
        {kingma2014adam}
\bibfield{author}{\bibinfo{person}{Diederik~P Kingma} {and}
  \bibinfo{person}{Jimmy Ba}.} \bibinfo{year}{2014}\natexlab{}.
\newblock \showarticletitle{Adam: A method for stochastic optimization}.
\newblock \bibinfo{journal}{\emph{arXiv preprint arXiv:1412.6980}}
  (\bibinfo{year}{2014}).
\newblock


\bibitem[\protect\citeauthoryear{Laine and Karras}{Laine and Karras}{2010}]%
        {laine2010efficient}
\bibfield{author}{\bibinfo{person}{Samuli Laine} {and} \bibinfo{person}{Tero
  Karras}.} \bibinfo{year}{2010}\natexlab{}.
\newblock \showarticletitle{Efficient sparse voxel octrees}.
\newblock \bibinfo{journal}{\emph{IEEE Transactions on Visualization and
  Computer Graphics}} \bibinfo{volume}{17}, \bibinfo{number}{8}
  (\bibinfo{year}{2010}), \bibinfo{pages}{1048--1059}.
\newblock


\bibitem[\protect\citeauthoryear{Liu, Gu, Lin, Chua, and Theobalt}{Liu
  et~al\mbox{.}}{2020}]%
        {liu2020neural}
\bibfield{author}{\bibinfo{person}{Lingjie Liu}, \bibinfo{person}{Jiatao Gu},
  \bibinfo{person}{Kyaw~Zaw Lin}, \bibinfo{person}{Tat-Seng Chua}, {and}
  \bibinfo{person}{Christian Theobalt}.} \bibinfo{year}{2020}\natexlab{}.
\newblock \showarticletitle{Neural Sparse Voxel Fields}.
\newblock \bibinfo{journal}{\emph{NeurIPS}} (\bibinfo{year}{2020}).
\newblock


\bibitem[\protect\citeauthoryear{Loviscach}{Loviscach}{2005}]%
        {loviscach2005efficient}
\bibfield{author}{\bibinfo{person}{J{\"o}rn Loviscach}.}
  \bibinfo{year}{2005}\natexlab{}.
\newblock \showarticletitle{Efficient magnification of bi-level textures}.
\newblock In \bibinfo{booktitle}{\emph{ACM SIGGRAPH 2005 Sketches}}.
  \bibinfo{pages}{131--es}.
\newblock


\bibitem[\protect\citeauthoryear{Martel, Lindell, Lin, Chan, Monteiro, and
  Wetzstein}{Martel et~al\mbox{.}}{2021}]%
        {martel2021acorn}
\bibfield{author}{\bibinfo{person}{Julien N.~P. Martel},
  \bibinfo{person}{David~B. Lindell}, \bibinfo{person}{Connor~Z. Lin},
  \bibinfo{person}{Eric~R. Chan}, \bibinfo{person}{Marco Monteiro}, {and}
  \bibinfo{person}{Gordon Wetzstein}.} \bibinfo{year}{2021}\natexlab{}.
\newblock \bibinfo{title}{ACORN: Adaptive Coordinate Networks for Neural Scene
  Representation}.
\newblock
\newblock
\showeprint[arxiv]{cs.CV/2105.02788}


\bibitem[\protect\citeauthoryear{Max}{Max}{1995}]%
        {max1995optical}
\bibfield{author}{\bibinfo{person}{Nelson Max}.}
  \bibinfo{year}{1995}\natexlab{}.
\newblock \showarticletitle{Optical models for direct volume rendering}.
\newblock \bibinfo{journal}{\emph{IEEE Transactions on Visualization and
  Computer Graphics}} \bibinfo{volume}{1}, \bibinfo{number}{2}
  (\bibinfo{year}{1995}), \bibinfo{pages}{99--108}.
\newblock


\bibitem[\protect\citeauthoryear{Mescheder, Oechsle, Niemeyer, Nowozin, and
  Geiger}{Mescheder et~al\mbox{.}}{2019}]%
        {mescheder2019occupancy}
\bibfield{author}{\bibinfo{person}{Lars Mescheder}, \bibinfo{person}{Michael
  Oechsle}, \bibinfo{person}{Michael Niemeyer}, \bibinfo{person}{Sebastian
  Nowozin}, {and} \bibinfo{person}{Andreas Geiger}.}
  \bibinfo{year}{2019}\natexlab{}.
\newblock \showarticletitle{Occupancy networks: Learning 3d reconstruction in
  function space}. In \bibinfo{booktitle}{\emph{Proceedings of the IEEE/CVF
  Conference on Computer Vision and Pattern Recognition}}.
  \bibinfo{pages}{4460--4470}.
\newblock


\bibitem[\protect\citeauthoryear{Mildenhall, Srinivasan, Tancik, Barron,
  Ramamoorthi, and Ng}{Mildenhall et~al\mbox{.}}{2020}]%
        {mildenhall2020nerf}
\bibfield{author}{\bibinfo{person}{Ben Mildenhall}, \bibinfo{person}{Pratul~P
  Srinivasan}, \bibinfo{person}{Matthew Tancik}, \bibinfo{person}{Jonathan~T
  Barron}, \bibinfo{person}{Ravi Ramamoorthi}, {and} \bibinfo{person}{Ren Ng}.}
  \bibinfo{year}{2020}\natexlab{}.
\newblock \showarticletitle{Nerf: Representing scenes as neural radiance fields
  for view synthesis}. In \bibinfo{booktitle}{\emph{ECCV}}.
  \bibinfo{pages}{405--421}.
\newblock


\bibitem[\protect\citeauthoryear{M\"uller, Evans, Schied, and Keller}{M\"uller
  et~al\mbox{.}}{2022}]%
        {mueller2022instant}
\bibfield{author}{\bibinfo{person}{Thomas M\"uller}, \bibinfo{person}{Alex
  Evans}, \bibinfo{person}{Christoph Schied}, {and} \bibinfo{person}{Alexander
  Keller}.} \bibinfo{year}{2022}\natexlab{}.
\newblock \showarticletitle{Instant Neural Graphics Primitives with a
  Multiresolution Hash Encoding}.
\newblock \bibinfo{journal}{\emph{arXiv:2201.05989}} (\bibinfo{date}{Jan.}
  \bibinfo{year}{2022}).
\newblock


\bibitem[\protect\citeauthoryear{Nguyen-Phuoc, Li, Theis, Richardt, and
  Yang}{Nguyen-Phuoc et~al\mbox{.}}{2019}]%
        {nguyen2019hologan}
\bibfield{author}{\bibinfo{person}{Thu Nguyen-Phuoc}, \bibinfo{person}{Chuan
  Li}, \bibinfo{person}{Lucas Theis}, \bibinfo{person}{Christian Richardt},
  {and} \bibinfo{person}{Yong-Liang Yang}.} \bibinfo{year}{2019}\natexlab{}.
\newblock \showarticletitle{Hologan: Unsupervised learning of 3d
  representations from natural images}. In \bibinfo{booktitle}{\emph{ICCV}}.
  \bibinfo{pages}{7588--7597}.
\newblock


\bibitem[\protect\citeauthoryear{Nguyen{-}Phuoc, Richardt, Mai, Yang, and
  Mitra}{Nguyen{-}Phuoc et~al\mbox{.}}{2020}]%
        {blockGAN:2020}
\bibfield{author}{\bibinfo{person}{Thu Nguyen{-}Phuoc},
  \bibinfo{person}{Christian Richardt}, \bibinfo{person}{Long Mai},
  \bibinfo{person}{Yong{-}Liang Yang}, {and} \bibinfo{person}{Niloy~J. Mitra}.}
  \bibinfo{year}{2020}\natexlab{}.
\newblock \showarticletitle{BlockGAN: Learning 3D Object-aware Scene
  Representations from Unlabelled Images}.
\newblock \bibinfo{journal}{\emph{CoRR}}  \bibinfo{volume}{abs/2002.08988}
  (\bibinfo{year}{2020}).
\newblock
\urldef\tempurl%
\url{https://arxiv.org/abs/2002.08988}
\showURL{%
\tempurl}


\bibitem[\protect\citeauthoryear{Niemeyer and Geiger}{Niemeyer and
  Geiger}{2021}]%
        {niemeyer2021giraffe}
\bibfield{author}{\bibinfo{person}{Michael Niemeyer} {and}
  \bibinfo{person}{Andreas Geiger}.} \bibinfo{year}{2021}\natexlab{}.
\newblock \showarticletitle{Giraffe: Representing scenes as compositional
  generative neural feature fields}. In \bibinfo{booktitle}{\emph{IEEE CVPR}}.
  \bibinfo{pages}{11453--11464}.
\newblock


\bibitem[\protect\citeauthoryear{Parilov and Zorin}{Parilov and Zorin}{2008}]%
        {parilov2008real}
\bibfield{author}{\bibinfo{person}{Evgueni Parilov} {and}
  \bibinfo{person}{Denis Zorin}.} \bibinfo{year}{2008}\natexlab{}.
\newblock \showarticletitle{Real-time rendering of textures with feature
  curves}.
\newblock \bibinfo{journal}{\emph{ACM Transactions on Graphics (TOG)}}
  \bibinfo{volume}{27}, \bibinfo{number}{1} (\bibinfo{year}{2008}),
  \bibinfo{pages}{1--15}.
\newblock


\bibitem[\protect\citeauthoryear{Park, Florence, Straub, Newcombe, and
  Lovegrove}{Park et~al\mbox{.}}{2019}]%
        {park2019deepsdf}
\bibfield{author}{\bibinfo{person}{Jeong~Joon Park}, \bibinfo{person}{Peter
  Florence}, \bibinfo{person}{Julian Straub}, \bibinfo{person}{Richard
  Newcombe}, {and} \bibinfo{person}{Steven Lovegrove}.}
  \bibinfo{year}{2019}\natexlab{}.
\newblock \showarticletitle{Deepsdf: Learning continuous signed distance
  functions for shape representation}. In \bibinfo{booktitle}{\emph{Proceedings
  of the IEEE/CVF Conference on Computer Vision and Pattern Recognition}}.
  \bibinfo{pages}{165--174}.
\newblock


\bibitem[\protect\citeauthoryear{Pavi{\'c} and Kobbelt}{Pavi{\'c} and
  Kobbelt}{2010}]%
        {pavic2010two}
\bibfield{author}{\bibinfo{person}{Darko Pavi{\'c}} {and} \bibinfo{person}{Leif
  Kobbelt}.} \bibinfo{year}{2010}\natexlab{}.
\newblock \showarticletitle{Two-Colored Pixels}. In
  \bibinfo{booktitle}{\emph{Computer Graphics Forum}},
  Vol.~\bibinfo{volume}{29}. Wiley Online Library, \bibinfo{pages}{743--752}.
\newblock


\bibitem[\protect\citeauthoryear{Ramanarayanan, Bala, and Walter}{Ramanarayanan
  et~al\mbox{.}}{2004}]%
        {ramanarayanan2004feature}
\bibfield{author}{\bibinfo{person}{Ganesh Ramanarayanan},
  \bibinfo{person}{Kavita Bala}, {and} \bibinfo{person}{Bruce Walter}.}
  \bibinfo{year}{2004}\natexlab{}.
\newblock \bibinfo{title}{Feature-based textures}.
\newblock
\newblock


\bibitem[\protect\citeauthoryear{Reiser, Peng, Liao, and Geiger}{Reiser
  et~al\mbox{.}}{2021}]%
        {reiser2021kilonerf}
\bibfield{author}{\bibinfo{person}{Christian Reiser}, \bibinfo{person}{Songyou
  Peng}, \bibinfo{person}{Yiyi Liao}, {and} \bibinfo{person}{Andreas Geiger}.}
  \bibinfo{year}{2021}\natexlab{}.
\newblock \showarticletitle{KiloNeRF: Speeding up Neural Radiance Fields with
  Thousands of Tiny MLPs}.
\newblock \bibinfo{journal}{\emph{arXiv preprint arXiv:2103.13744}}
  (\bibinfo{year}{2021}).
\newblock


\bibitem[\protect\citeauthoryear{Schonberger and Frahm}{Schonberger and
  Frahm}{2016}]%
        {schonberger2016structure}
\bibfield{author}{\bibinfo{person}{Johannes~L Schonberger} {and}
  \bibinfo{person}{Jan-Michael Frahm}.} \bibinfo{year}{2016}\natexlab{}.
\newblock \showarticletitle{Structure-from-motion revisited}. In
  \bibinfo{booktitle}{\emph{IEEE CVPR}}. \bibinfo{pages}{4104--4113}.
\newblock


\bibitem[\protect\citeauthoryear{Schwarz, Liao, Niemeyer, and Geiger}{Schwarz
  et~al\mbox{.}}{2020}]%
        {schwarz2020graf}
\bibfield{author}{\bibinfo{person}{Katja Schwarz}, \bibinfo{person}{Yiyi Liao},
  \bibinfo{person}{Michael Niemeyer}, {and} \bibinfo{person}{Andreas Geiger}.}
  \bibinfo{year}{2020}\natexlab{}.
\newblock \showarticletitle{Graf: Generative radiance fields for 3d-aware image
  synthesis}.
\newblock \bibinfo{journal}{\emph{arXiv preprint arXiv:2007.02442}}
  (\bibinfo{year}{2020}).
\newblock


\bibitem[\protect\citeauthoryear{Sen}{Sen}{2004}]%
        {sen2004silhouette}
\bibfield{author}{\bibinfo{person}{Pradeep Sen}.}
  \bibinfo{year}{2004}\natexlab{}.
\newblock \showarticletitle{Silhouette maps for improved texture
  magnification}. In \bibinfo{booktitle}{\emph{Proceedings of the ACM
  SIGGRAPH/EUROGRAPHICS conference on Graphics hardware}}.
  \bibinfo{pages}{65--73}.
\newblock


\bibitem[\protect\citeauthoryear{Sen, Cammarano, and Hanrahan}{Sen
  et~al\mbox{.}}{2003}]%
        {sen2003shadow}
\bibfield{author}{\bibinfo{person}{Pradeep Sen}, \bibinfo{person}{Mike
  Cammarano}, {and} \bibinfo{person}{Pat Hanrahan}.}
  \bibinfo{year}{2003}\natexlab{}.
\newblock \showarticletitle{Shadow silhouette maps}.
\newblock \bibinfo{journal}{\emph{ACM Transactions on Graphics (TOG)}}
  \bibinfo{volume}{22}, \bibinfo{number}{3} (\bibinfo{year}{2003}),
  \bibinfo{pages}{521--526}.
\newblock


\bibitem[\protect\citeauthoryear{Sitzmann, Martel, Bergman, Lindell, and
  Wetzstein}{Sitzmann et~al\mbox{.}}{2020}]%
        {sitzmann2020implicit}
\bibfield{author}{\bibinfo{person}{Vincent Sitzmann}, \bibinfo{person}{Julien
  N.~P. Martel}, \bibinfo{person}{Alexander~W. Bergman},
  \bibinfo{person}{David~B. Lindell}, {and} \bibinfo{person}{Gordon
  Wetzstein}.} \bibinfo{year}{2020}\natexlab{}.
\newblock \bibinfo{title}{Implicit Neural Representations with Periodic
  Activation Functions}.
\newblock
\newblock
\showeprint[arxiv]{cs.CV/2006.09661}


\bibitem[\protect\citeauthoryear{Sitzmann, Thies, Heide, Nießner, Wetzstein,
  and Zollhöfer}{Sitzmann et~al\mbox{.}}{2019}]%
        {sitzmann2019deepvoxels}
\bibfield{author}{\bibinfo{person}{Vincent Sitzmann}, \bibinfo{person}{Justus
  Thies}, \bibinfo{person}{Felix Heide}, \bibinfo{person}{Matthias Nießner},
  \bibinfo{person}{Gordon Wetzstein}, {and} \bibinfo{person}{Michael
  Zollhöfer}.} \bibinfo{year}{2019}\natexlab{}.
\newblock \bibinfo{title}{DeepVoxels: Learning Persistent 3D Feature
  Embeddings}.
\newblock
\newblock
\showeprint[arxiv]{cs.CV/1812.01024}


\bibitem[\protect\citeauthoryear{Sun, Sun, and Chen}{Sun et~al\mbox{.}}{2021}]%
        {sun2021direct}
\bibfield{author}{\bibinfo{person}{Cheng Sun}, \bibinfo{person}{Min Sun}, {and}
  \bibinfo{person}{Hwann-Tzong Chen}.} \bibinfo{year}{2021}\natexlab{}.
\newblock \bibinfo{title}{Direct Voxel Grid Optimization: Super-fast
  Convergence for Radiance Fields Reconstruction}.
\newblock
\newblock
\showeprint[arxiv]{cs.CV/2111.11215}


\bibitem[\protect\citeauthoryear{Tarini and Cignoni}{Tarini and
  Cignoni}{2005}]%
        {tarini2005pinchmaps}
\bibfield{author}{\bibinfo{person}{Marco Tarini} {and} \bibinfo{person}{Paolo
  Cignoni}.} \bibinfo{year}{2005}\natexlab{}.
\newblock \showarticletitle{Pinchmaps: Textures with customizable
  discontinuities}. In \bibinfo{booktitle}{\emph{Computer Graphics Forum}},
  Vol.~\bibinfo{volume}{24}. Blackwell Publishing, Inc Oxford, UK and Boston,
  USA, \bibinfo{pages}{557--568}.
\newblock


\bibitem[\protect\citeauthoryear{Tulsiani, Zhou, Efros, and Malik}{Tulsiani
  et~al\mbox{.}}{2017}]%
        {tulsiani2017multi}
\bibfield{author}{\bibinfo{person}{Shubham Tulsiani}, \bibinfo{person}{Tinghui
  Zhou}, \bibinfo{person}{Alexei~A Efros}, {and} \bibinfo{person}{Jitendra
  Malik}.} \bibinfo{year}{2017}\natexlab{}.
\newblock \showarticletitle{Multi-view supervision for single-view
  reconstruction via differentiable ray consistency}. In
  \bibinfo{booktitle}{\emph{Proceedings of the IEEE conference on computer
  vision and pattern recognition}}. \bibinfo{pages}{2626--2634}.
\newblock


\bibitem[\protect\citeauthoryear{Tumblin and Choudhury}{Tumblin and
  Choudhury}{2004}]%
        {tumblin2004bixels}
\bibfield{author}{\bibinfo{person}{Jack Tumblin} {and} \bibinfo{person}{Prasun
  Choudhury}.} \bibinfo{year}{2004}\natexlab{}.
\newblock \showarticletitle{Bixels: Picture samples with sharp embedded
  boundaries.}. In \bibinfo{booktitle}{\emph{Rendering Techniques}}. Citeseer,
  \bibinfo{pages}{255--264}.
\newblock


\bibitem[\protect\citeauthoryear{Wizadwongsa, Phongthawee, Yenphraphai, and
  Suwajanakorn}{Wizadwongsa et~al\mbox{.}}{2021}]%
        {wizadwongsa2021nex}
\bibfield{author}{\bibinfo{person}{Suttisak Wizadwongsa},
  \bibinfo{person}{Pakkapon Phongthawee}, \bibinfo{person}{Jiraphon
  Yenphraphai}, {and} \bibinfo{person}{Supasorn Suwajanakorn}.}
  \bibinfo{year}{2021}\natexlab{}.
\newblock \showarticletitle{Nex: Real-time view synthesis with neural basis
  expansion}. In \bibinfo{booktitle}{\emph{Proceedings of the IEEE/CVF
  Conference on Computer Vision and Pattern Recognition}}.
  \bibinfo{pages}{8534--8543}.
\newblock


\bibitem[\protect\citeauthoryear{Yu, Fridovich-Keil, Tancik, Chen, Recht, and
  Kanazawa}{Yu et~al\mbox{.}}{2021a}]%
        {yu2021plenoxels}
\bibfield{author}{\bibinfo{person}{Alex Yu}, \bibinfo{person}{Sara
  Fridovich-Keil}, \bibinfo{person}{Matthew Tancik}, \bibinfo{person}{Qinhong
  Chen}, \bibinfo{person}{Benjamin Recht}, {and} \bibinfo{person}{Angjoo
  Kanazawa}.} \bibinfo{year}{2021}\natexlab{a}.
\newblock \bibinfo{title}{Plenoxels: Radiance Fields without Neural Networks}.
\newblock
\newblock
\showeprint[arxiv]{cs.CV/2112.05131}


\bibitem[\protect\citeauthoryear{Yu, Li, Tancik, Li, Ng, and Kanazawa}{Yu
  et~al\mbox{.}}{2021b}]%
        {yu2021plenoctrees}
\bibfield{author}{\bibinfo{person}{Alex Yu}, \bibinfo{person}{Ruilong Li},
  \bibinfo{person}{Matthew Tancik}, \bibinfo{person}{Hao Li},
  \bibinfo{person}{Ren Ng}, {and} \bibinfo{person}{Angjoo Kanazawa}.}
  \bibinfo{year}{2021}\natexlab{b}.
\newblock \showarticletitle{Plenoctrees for real-time rendering of neural
  radiance fields}.
\newblock \bibinfo{journal}{\emph{arXiv preprint arXiv:2103.14024}}
  (\bibinfo{year}{2021}).
\newblock


\bibitem[\protect\citeauthoryear{Zhang, Isola, Efros, Shechtman, and
  Wang}{Zhang et~al\mbox{.}}{2018}]%
        {zhang2018unreasonable}
\bibfield{author}{\bibinfo{person}{Richard Zhang}, \bibinfo{person}{Phillip
  Isola}, \bibinfo{person}{Alexei~A Efros}, \bibinfo{person}{Eli Shechtman},
  {and} \bibinfo{person}{Oliver Wang}.} \bibinfo{year}{2018}\natexlab{}.
\newblock \showarticletitle{The unreasonable effectiveness of deep features as
  a perceptual metric}. In \bibinfo{booktitle}{\emph{Proceedings of the IEEE
  conference on computer vision and pattern recognition}}.
  \bibinfo{pages}{586--595}.
\newblock


\end{thebibliography}

\end{document}